\pdfoutput=1

\documentclass[11pt]{article}

\usepackage[final]{acl}

\usepackage{times}
\usepackage{latexsym}
\usepackage{amsfonts}
\usepackage{arydshln}
\usepackage{balance}

\usepackage[T1]{fontenc}

\usepackage[utf8]{inputenc}

\usepackage{microtype}
\usepackage{comment}
\usepackage[ruled,vlined]{algorithm2e}
\usepackage{amsmath}
\usepackage{inconsolata}
\usepackage{multirow}
\usepackage{makecell}
\usepackage{colortbl}
\usepackage{graphicx}
\usepackage{booktabs}
\usepackage{subfig}
\usepackage{pifont}
\usepackage{enumitem}
\usepackage{algpseudocode}
\usepackage{soul}
\usepackage[most]{tcolorbox}
\tcbuselibrary{listings}
\tcbuselibrary{skins}
\tcbuselibrary{breakable}
\usepackage{bbding}
\usepackage{scalerel}
\usepackage{colortbl}
  \usepackage{tcolorbox}
  \usepackage{multicol}
  \tcbuselibrary{listings, breakable, skins}
\title{Agentic Chain-of-Thought Steering for Efficient and Controllable \\ LLM Reasoning}

\author{Yu Xia$^{1}$\quad Zhouhang Xie$^{1}$\quad Xin Xu$^{1}$\quad Byungkyu Kang$^{2}$ \\
{\bf Prarit Lamba$^{2}$\quad Xiang Gao$^{2}$\quad Julian McAuley$^{1}$} \\
   University of California San Diego$^{1}$ \quad Intuit$^{2}$ \\
    \texttt{\{yux078, zhx022, xinxucs, jmcauley\}@ucsd.edu} \\
    \texttt{\{Jay\_Kang, Prarit\_Lamba, Xiang\_Gao\}@intuit.com}}

\begin{document}
\maketitle
\begin{abstract}
    Large language models improve final-answer accuracy through extended chain-of-thought reasoning, but often spend tokens inefficiently and offer little inference-time control. Existing efficient reasoning methods control thinking length by shortening, early-stopping, or compressing traces, leaving how the model thinks implicit. In this paper, we propose Agentic Chain-of-Thought Steering (ACTS), which formulates reasoning steering as a Markov decision process where a controller agent adaptively steers a frozen reasoner during inference. At each step, the controller observes the reasoning trace and remaining thinking budget, then issues a steering action consisting of a reasoning strategy and a steering phrase that initiates the next reasoner step. This enables budget-aware strategy control for efficient reasoning while preserving the reasoner's generation continuity. We initialize the controller agent from our constructed synthetic steering trajectories with multi-budget augmentation, and further optimize it via reinforcement learning with budget-conditioned reward shaping. Experiments across multiple benchmarks show that ACTS achieves competitive accuracy with substantial token savings, and enables controllable accuracy--token trade-offs across different reasoners and tasks. The code is available at \url{https://github.com/Andree-9/ACTS}.
\end{abstract}

\section{Introduction}

Large language models (LLMs) improve final answer accuracy by spending test-time computation on extended chain-of-thought reasoning~\citep{wei2022chain, xia-etal-2025-beyond, guo2025deepseek}. However, long reasoning traces often contain redundant derivations, repeated self-checks, and delayed termination, making reasoning expensive and difficult to control \cite{feng2025efficient,xia-etal-2025-selection,sui2025stop}, adding to broader concerns over efficient LLM deployment~\cite{nguyen-etal-2025-survey,xia2024which}. Recent efficient reasoning methods reduce this cost by shortening~\cite{aggarwal2025l1, hou2025thinkprune}, compressing~\cite{xia-etal-2025-tokenskip, zhang-etal-2025-lightthinker,ThinkRouter}, early-stopping~\cite{yang2026dynamic, liu-wang-2025-answer}, or budget-forcing~\cite{muennighoff-etal-2025-s1, wen2025budgetthinker} the generated chain-of-thought thinking trace.

While effective, these methods mainly control how long the model thinks, leaving how it thinks largely implicit. A reasoning trace is not a uniform token sequence: it consists of functional steps such as setting up the problem, carrying out derivations, switching to alternatives, verifying results, and deciding when to stop. Recent studies identify structured transitions between such behaviors, e.g., planning, execution, reflection, and conclusion, in long reasoning traces~\citep{li-etal-2025-understanding,xiong-etal-2025-mapping, li-etal-2026-thinkpilot}. This motivates a different control surface for efficient reasoning: steering the stepwise reasoning behavior itself, rather than imposing only a global length constraint.

In this paper, we propose Agentic Chain-of-Thought Steering (ACTS), a framework for efficient and controllable LLM reasoning. We formulate reasoning steering under budget as a Markov decision process in which a controller agent guides a frozen reasoner step by step. At each step, the controller observes the steering history, current reasoning step, and remaining thinking budget, then selects a steering action consisting of a high-level reasoning strategy and a short natural-language steering phrase that initiates the next reasoner step. This design enables reasoner-agnostic in-flight strategy control while preserving the continuity of the reasoner's thinking process.

\begin{figure*}[t]
      \centering
      \includegraphics[width=\linewidth]{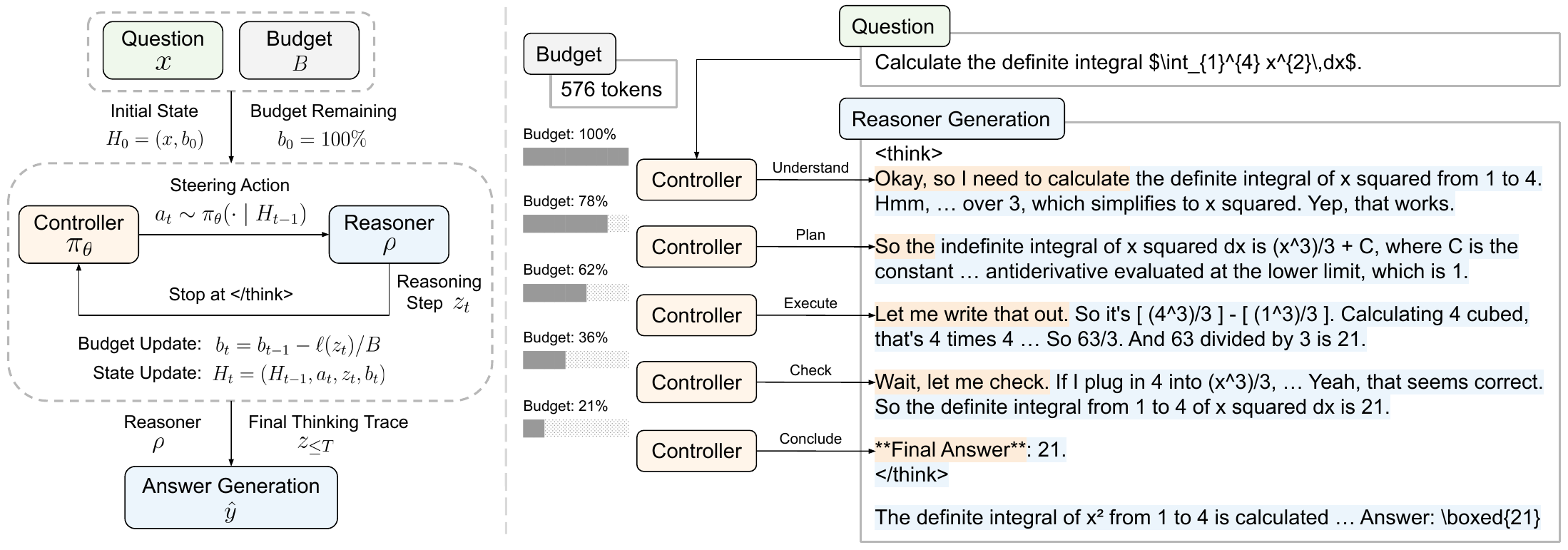}
      \caption{Overview of \textsc{ACTS}. {Left:} a controller agent steers a frozen reasoner step by step under a
  thinking-token budget (Detailed formulation in Section \ref{sec:mdp}). {Right:} an illustrative example of controller-steered reasoner generation.}
      \label{fig:overview}
\end{figure*}

To initialize the controller, we need expert steering trajectories that specify the desired strategy and steering phrase at each reasoning step. Existing expert traces, however, contain only the reasoner's chain of thought, without explicit controller actions. We therefore synthesize steering trajectories by segmenting each trace, annotating the implied strategy, and extracting the step-opening phrase as the steering phrase. We further apply multi-budget augmentation to expose the controller to varying budget regimes, and train it with supervised fine-tuning.
We then optimize the controller via reinforcement learning with budget-conditioned reward shaping. Specifically, we penalize two scenarios: (i) producing an answer while overusing the budget, which discourages overthinking, and (ii) terminating with an incorrect answer while underusing the budget, which discourages premature termination on hard questions.
At inference time, we instantiate the controller and reasoner as decoupled asynchronous servers, achieving aggregate throughput close to direct reasoner generation despite the added controller calls.
In summary, we make the following contributions:
\begin{itemize}[leftmargin=*, itemsep=2pt]
    \item We propose Agentic Chain-of-Thought Steering (ACTS), a reasoner-agnostic framework that treats efficient reasoning as strategy-level steering over an evolving chain of thought, rather than global length control.
    \item We train the controller agent with two stages: behavior initialization from our constructed synthetic steering trajectories with multi-budget augmentation, followed by reinforcement learning with budget-conditioned reward shaping.
    \item Experiments across multiple benchmarks show that ACTS achieves competitive accuracy under substantially lower token usage, and enables controllable accuracy--token trade-offs across different reasoners and tasks.
\end{itemize}

\section{Related Work}

\subsection{Efficient and Budget-Aware Reasoning}

A growing body of work seeks to reduce the cost of long chain-of-thought reasoning. Prompt-level approaches encourage concise traces via brevity instructions, token-budget prompts, or minimal informative drafts~\cite{nayab2024concise, han-etal-2025-token, xu2025chain}. Tuning-based methods learn length-controllable variants of CoT through supervised compression of reasoning traces~\cite{xia-etal-2025-tokenskip, ma-etal-2025-cot, zhang-etal-2025-lightthinker}, while RL-based methods train reasoners to satisfy explicit length constraints with length-aware rewards~\cite{luo2025o1, hou2025thinkprune, aggarwal2025l1, he2025smartthinker}. Complementary inference-time methods directly truncate or guide generation via budget forcing or budget-aware control signals~\cite{muennighoff-etal-2025-s1, li2025steering, wen2025budgetthinker}, or via early-exit once the model is confident in a trial answer~\cite{yang2026dynamic, liu-wang-2025-answer,jiang2025flashthink}. While effective, these methods primarily control how long the model thinks, leaving the underlying reasoning strategy implicit. ACTS is complementary: it steers how the model thinks at each step through explicit reasoning strategies and steering phrases, with budget compliance arising from the learned completion rather than enforced length.

\subsection{Reasoning Steering and Thinking Control}

Beyond length, a separate line of work intervenes directly in the reasoning process to control its behavior. Prompt-level steering inserts strategic instructions before the thinking trace to elicit specific reasoning behaviors~\cite{li-etal-2026-thinkpilot}. Activation steering extracts intervention vectors in the model's hidden representations to amplify or suppress cognitive behaviors such as reflection and backtracking at inference time~\cite{chen2025seal, zhang2025understanding, venhoff2025understanding, wang2026two}. A third thread trains explicit controller models that select reasoning structures or strategies at inference time via reinforcement learning~\cite{hao2025rlot, xia-etal-2025-sand,xia2026learning,wu2025ocean, liu-etal-2025-epo}. 
Closely related, meta-cognitive frameworks place a higher-level reasoner above the generator to adapt strategies during inference~\cite{xiang2025towards,shen2026simultaneous, sui2025meta}. Analyses also identify recurring functional structure in long reasoning traces~\cite{li-etal-2025-understanding, xiong-etal-2025-mapping}, motivating control surfaces aligned with these observed behaviors. ACTS unifies these threads: it learns a reasoner-agnostic controller agent that emits an explicit strategy and a free-form steering phrase at each step, optimized via reinforcement learning with budget-conditioned reward shaping that jointly captures answer correctness and budget utilization.

\section{Agentic Chain-of-Thought Steering}

In this section, we first formalize the reasoning steering task in Section~\ref{sec:mdp}, then describe the two-stage controller agent training in Sections~\ref{sec:sft} and~\ref{sec:rl}. Section~\ref{sec:inference} presents the asynchronous controller-reasoner inference loop. Figure~\ref{fig:overview} provides an overview of the ACTS framework.

\subsection{Reasoning Steering as an MDP}
\label{sec:mdp}

We formulate reasoning steering under budget as a Markov decision process (MDP). Given a question $x \in \mathcal{X}$ and a thinking-token budget $B \in \mathbb{N}^{+}$, the goal is to produce a final answer $\hat{y}$ via controller-steered reasoner generation. We denote the controller agent policy by $\pi_\theta$ and the frozen reasoner by $\rho$, and initialize the steering history as $H_0=(x,b_0)$, where $b_t$ is the fraction of the budget remaining at step $t$ and $b_0=100\%$.
Following prior work on chain-of-thought segmentation~\cite{yang2025speculative, ling2026neural}, we define a reasoning step as a self-contained semantic unit such as calculations and verifications delimited by paragraph boundaries, e.g., \texttt{".\textbackslash n\textbackslash n"} and \texttt{"?\textbackslash n\textbackslash n"}. We write the $t$-th reasoner step as $z_t$ and the thinking trace through step $t$ as $z_{\leq t}=z_1 \circ \cdots \circ z_t$.

\paragraph{State and Action.}
At step $t$, the controller observes the full steering history $H_{t-1}$ as the state, which records previous controller actions, reasoner steps, and remaining budgets. It samples a steering action $a_t=(u_t,p_t) \sim \pi_\theta(\cdot \mid H_{t-1})$, where $u_t \in \mathcal{U}$ is a high-level reasoning strategy from a predefined set and $p_t$ is a free-form natural-language steering phrase that primes the reasoner's next step toward $u_t$. We follow taxonomies of reasoning behavior~\cite{schoenfeld2014mathematical,li-etal-2025-understanding,li-etal-2026-thinkpilot} and instantiate the strategy set as
$\mathcal{U} = \{$\textsc{understand}, \textsc{plan}, \textsc{execute}, \textsc{explore}, \textsc{check}, \textsc{summarize}, \textsc{conclude}$\}$.
The steering phrase $p_t$ is a short discourse phrase learned by the controller agent; for example, \emph{``Wait, let me verify''} for \textsc{check} or \emph{``Alternatively, suppose''} for \textsc{explore}. This decouples what strategy to take from how to enter it linguistically, preserving continuity with the reasoner's natural generation style.
A more detailed instantiation is provided in Appendix \ref{app:strategies}.

\paragraph{Transition.}
Conditioned on the question $x$, the current thinking trace $z_{<t}$, and the controller-emitted steering phrase $p_t$ that enacts the selected strategy $u_t$, the reasoner generates a continuation $s_t \sim \rho(\cdot \mid x,z_{<t},p_t)$, yielding the next reasoner step $z_t = p_t \circ s_t$. We then update the remaining-budget fraction $b_t = b_{t-1} - \ell(z_t)/B$ and steering history $H_t = (H_{t-1}, a_t, z_t, b_t)$, where $\ell(\cdot)$ counts thinking tokens. The episode terminates when the controller selects \textsc{conclude}, the reasoner emits an end-of-thinking token (e.g., \texttt{</think>}), or a maximum step count is reached.

\paragraph{Terminal Reward.}
After the thinking episode terminates at step $T$, we run an answer-generation step with the reasoner conditioned on the completed thinking trace $z_{\leq T}$ and extract the final answer $\hat{y}$. We denote the resulting steering trajectory as
$\tau = (x,\, b_0,\, a_1, z_1, b_1,\, \ldots,\, a_T, z_T, b_T)$.
Given the ground-truth answer $y^{\star}$, the terminal reward is defined as a function of answer correctness $c = \mathbf{1}[\hat{y}=y^{\star}]$ and the remaining-budget fraction $b_T$ at termination,
\begin{equation}\label{eq:reward}
    R(\tau,\hat{y}) = f(c,\, b_T)\,.
\end{equation}
We specify our concrete budget-conditioned reward shaping in Section~\ref{sec:rl}.

\begin{figure}[t]
      \centering
      \includegraphics[width=\linewidth]{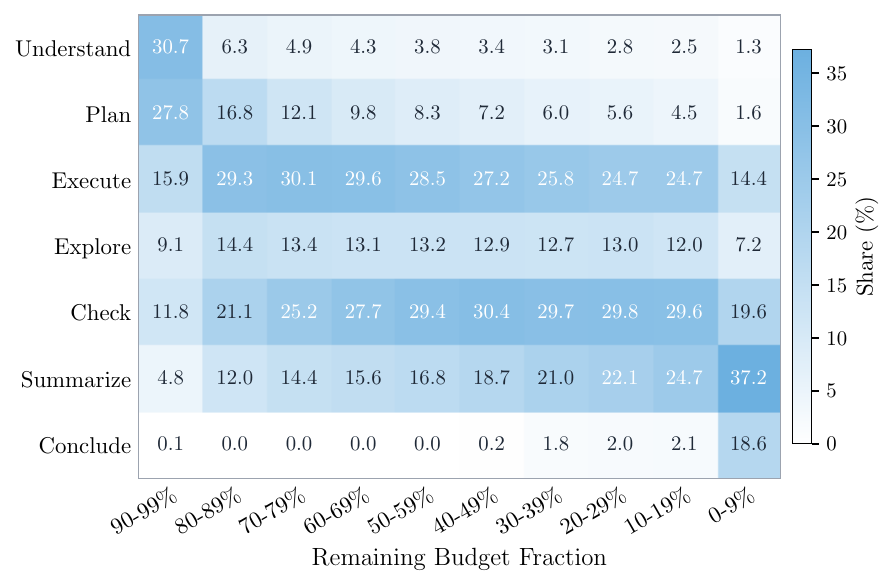}
      \caption{Joint distribution of controller strategy and remaining budget over the synthetic steering corpus.}
      \label{fig:heatmap-sft-data}
\end{figure}

\subsection{Controller Behavior Initialization}
\label{sec:sft}

To initialize the controller agent $\pi_\theta$, we construct a corpus of synthetic steering trajectories by extracting the controller action sequence implicit in expert reasoning traces, then fit $\pi_\theta$ to it under a standard supervised objective.

\paragraph{Synthetic Trajectory Construction.}
We build the corpus from the OpenR1-Math dataset~\cite{openr1} with reasoning traces produced by DeepSeek-R1~\cite{guo2025deepseek}. Each trace is segmented into reasoning steps by the paragraph boundaries defined in Section \ref{sec:mdp}. For every step $z_t$, we prompt an LLM annotator to classify the underlying reasoning strategy $u_t \in \mathcal{U}$ and extract the natural-language steering phrase $p_t$ that opens the step, yielding the controller action $a_t=(u_t,p_t)$. We simulate the per-step budget signal by taking the trace's own thinking-token length as the synthetic budget $B$ and consuming $\ell(z_s)$ at each step, so that $b_t = 1 - \sum_{s \le t}\ell(z_s)/B$. 
With a sequence of $(b_t, a_t, z_t)$ tuples, each trace is thereby assembled into a steering trajectory $\tau$.
We filter malformed annotations and trajectories containing degenerate looping patterns.
More trajectory construction details are provided in Appendix~\ref{app:sft}.

\paragraph{Multi-Budget Augmentation.}
Trajectories constructed with the synthetic budget $B$ always end exactly at budget exhaustion, exposing the controller to only a narrow termination regime. To improve the controller's budget awareness and flexibility, we augment each trajectory by rescaling its per-step budgets as follows so that it terminates at varying remaining budgets:
\begin{itemize}[leftmargin=*, itemsep=2pt]
    \item \textbf{Exhausted budget} ($b_T = 0\%$): trains the controller to conclude decisively at exhaustion.
    \item \textbf{Early termination} ($0\% < b_T \le 40\%$): trains the controller to conclude once the reasoner can finalize the answer, without using all budget.
\end{itemize}
The same expert trace thus appears across multiple budget scenarios, simulating the different user-specified budgets and question difficulties encountered at deployment.

\paragraph{Reasoning Strategy Curriculum.}
Figure~\ref{fig:heatmap-sft-data} plots the joint distribution of strategy $u_t$ and remaining-budget fraction $b_t$ over the corpus. Although the expert reasoner produces each trace without any budget conditioning, mapping token positions onto our synthetic budget axis exposes a clear temporal structure: \textsc{understand} and \textsc{plan} concentrate at the trace opening, \textsc{execute} holds a broad middle band, \textsc{check} rises through the mid-to-late range, and \textsc{summarize} and \textsc{conclude} dominate as the budget exhaustion. By anchoring the budget axis to trace position during steering trajectory construction, we turn this natural reasoning progression into a budget-conditioned distribution. Behavior initialization then transfers this distribution into the controller policy $\pi_\theta$ as a prior over when each strategy should fire under a given budget.

\begin{figure}[t]
      \centering
      \includegraphics[width=0.85\linewidth]{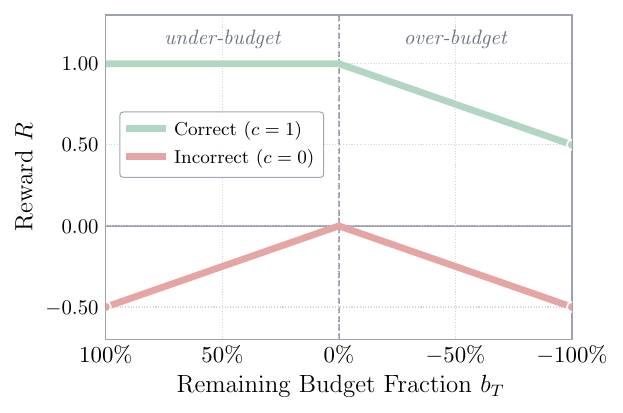}
      \caption{Budget-conditioned reward shaping.}
      \label{fig:reward_curve}
\end{figure}

\paragraph{Supervised Finetuning Objective.}
Let $\mathcal{D}=\{\tau^{(i)}\}_i$ denote the resulting set of synthetic steering trajectories. We initialize the controller agent $\pi_\theta$ by minimizing the loss on each controller turn,
\begin{equation}
    \mathcal{L}_{\text{SFT}}(\pi_\theta) = -\mathbb{E}_{\tau \sim \mathcal{D}}\sum_{t=1}^{T} \log \pi_\theta\!\left( a_t \mid H_{t-1} \right)\,,
\end{equation}
where $a_t = (u_t, p_t)$ is the controller turn and $H_{t-1}$ is the steering history through step $t-1$.

\subsection{Online Reinforcement Learning}
\label{sec:rl}
To optimize for answer accuracy and budget compliance, we further train the controller agent with reinforcement learning. 
However, simply applying an over-budget penalty to the correctness reward creates a perverse incentive: the controller can avoid the penalty by terminating early on questions it has not yet solved, trading correctness for budget compliance. 
Therefore, we instantiate $f(c, b_T)$ with budget-conditioned reward shaping that penalizes both overthinking and premature termination.

\paragraph{Budget-Conditioned Reward Shaping.}
We design a reward that targets two scenarios: (i) producing an answer while overspending the budget, signaling overthinking, and (ii) terminating with an incorrect answer while leaving budget underutilized, signaling premature termination on an unsolved question. Recall that the remaining-budget fraction $b_T$ is positive when budget is underused and negative when over-budget. With the answer correctness $c = \mathbf{1}[\hat{y}=y^\star]$, the shaped reward is then given as
\begin{equation}\label{eq:reward-shape}
    R(\tau, \hat{y}) =
    \begin{cases}
        1 + \alpha\, \min(b_T,\, 0), & c = 1, \\[2pt]
        -\alpha\, |b_T|, & c = 0,
    \end{cases}
\end{equation}
where $\alpha \in [0,1]$ controls the penalty magnitude. As shown in Figure~\ref{fig:reward_curve}, the reward for a correct answer decays from $1$ to $1-\alpha$ as the over-budget excess grows, while the reward for an incorrect answer decays from $0$ to $-\alpha$ as either the over-budget excess or the underused budget grows.
In practice, we additionally apply a $10\%$ grace margin around $b_T = 0$ so that minor over- or under-shoots near the budget boundary do not trigger the penalty.

\paragraph{Online GRPO Update.}
We optimize the controller $\pi_\theta$ over full steering trajectories with Group Relative Policy Optimization~\cite{shao2024deepseekmath}. For each question $x$, we sample $G$ trajectories $\{\tau_i\}_{i=1}^{G}$ by rolling out $\pi_\theta$ jointly with a frozen reasoner $\rho$, score each with the shaped reward $R_i$ from Equation~\eqref{eq:reward-shape}. Following Dr.\ GRPO~\cite{liu2025drgrpo}, we drop the standard-deviation normalization that introduces length and difficulty biases, and use the mean-centered group-relative advantage
\begin{equation}
    A_i = R_i - \bar{R}, \quad \bar{R} = \frac{1}{G}\sum_{i=1}^{G} R_i.
\end{equation}
The trajectory-level advantage $A_i$ is broadcast to all controller action tokens in $\tau_i$, while reasoner-generated continuation tokens at each step are masked out of the policy gradient. 

\subsection{Asynchronous Inference-Time Steering}
\label{sec:inference}

While ACTS enables adaptive reasoning steering, the controller-reasoner architecture adds serving overhead at inference-time deployment of our trained controller agent. 
Specifically, each sample alternates calls between the controller agent $\pi_\theta$ and the reasoner $\rho$ over many turns, and serializing these calls naively would inflate per-sample latency and degrade rollout throughput.
We address this with an asynchronous two-server pipeline, where the controller agent and the reasoner are hosted as separate SGLang~\cite{zheng2024sglang} servers, and the per-sample MDP loop is driven by an orchestrator over asynchronous HTTP. 
Since the two servers operate independently, all in-flight samples advance concurrently at request level, amortizing per-call latency while maintaining aggregate throughput. 
We empirically show in Section~\ref{sec:latency-analysis} that our async pipeline achieves aggregate throughput close to direct reasoner batched generation despite the additional controller calls.

\section{Experimental Setup}\label{sec:setup}

\subsection{Datasets and Evaluation}
\label{sec:setup-data}

As detailed in Section~\ref{sec:sft}, we construct synthetic steering trajectories for controller agent SFT from OpenR1-Math corpus~\cite{openr1}, from which we sample $7{,}500$ questions whose expert thinking traces are between $512$ and $8{,}192$ tokens long. The same question set is reused for online RL in Section~\ref{sec:rl}.
We evaluate our ACTS framework on four math reasoning benchmarks MATH-500~\cite{hendrycks2021measuring}, AIME24~\cite{maa2024aime}, AMC \cite{li2024numinamath} including both AMC12 2022 and 2023, and the math subset of OlympiadBench~\cite{he-etal-2024-olympiadbench}, and also a science QA benchmark GPQA Diamond~\cite{rein2024gpqa} to test whether our trained controller agent can generalize to non-math domains.
We report {Accuracy} by answer correctness and total {\#Tokens} including both thinking and answer generation counted by the reasoner tokenizer as main metrics. 
For our ACTS, {\#Tokens} includes both controller and reasoner tokens.
We run multiple seeded repeats and report per-sample averages on small-size benchmarks following \cite{yang2026dynamic}: $5$ repeats on AIME24 and AMC, and $3$ repeats on GPQA Diamond.
The budgets used for Table~\ref{tab:main-results} are reported in Appendix~\ref{app:additional-results}.
We use three reasoning models DeepSeek-R1-Distill-Qwen-$\{$1.5B,7B$\}$ \cite{guo2025deepseek}, and Qwen3-8B \cite{yang2025qwen3} as the reasoners in our evaluation.

\begin{table*}[t]
    \centering
    \resizebox{\textwidth}{!}{%
    \begin{tabular}{l ccc ccc ccc ccc ccc}
    \toprule
    \multirow{2}{*}{\textbf{Method}} & \multicolumn{3}{c}{\textbf{MATH-500}} & \multicolumn{3}{c}{\textbf{AIME24}} & \multicolumn{3}{c}{\textbf{AMC}} & \multicolumn{3}{c}{\textbf{OlympiadBench}} & \multicolumn{3}{c}{\textbf{GPQA Diamond}} \\
    \cmidrule(lr){2-4} \cmidrule(lr){5-7} \cmidrule(lr){8-10} \cmidrule(lr){11-13} \cmidrule(lr){14-16}
     & Acc~$\uparrow$ & \#Tok~$\downarrow$ & $\Delta\%$ & Acc~$\uparrow$ & \#Tok~$\downarrow$ & $\Delta\%$ & Acc~$\uparrow$ & \#Tok~$\downarrow$ & $\Delta\%$ & Acc~$\uparrow$ & \#Tok~$\downarrow$ & $\Delta\%$ & Acc~$\uparrow$ & \#Tok~$\downarrow$ & $\Delta\%$ \\
    \midrule
    \multicolumn{16}{l}{\textit{DeepSeek-R1-Distill-Qwen-1.5B}} \\
    \midrule
    Vanilla       & 83.6 & 5{,}428  & -     & 28.0 & 16{,}776 & -     & 63.1 & 10{,}835 & -     & 42.5 & 12{,}418 & -     & 18.2 & 9{,}285  & -     \\
    NoThink    & 70.0 & 1{,}246  & $-77.0$ & 14.0 & 5{,}679  & $-66.2$ & 49.2 & 2{,}394  & $-77.9$ & 38.1 & 2{,}162  & $-82.6$ & 19.2 & 977      & $-89.5$ \\
    CoD           & 60.2 & 3{,}832  & $-29.4$ & 26.7 & 12{,}891 & $-23.2$ & 53.5 & 7{,}421  & $-31.5$ & 33.9 & 8{,}499  & $-31.6$ & 20.9 & 7{,}993  & $-13.9$ \\
    DEER          & 63.8 & 2{,}073  & $-61.8$ & 30.7 & 8{,}392  & $-50.0$ & 54.0 & 5{,}169  & $-52.3$ & 33.5 & 4{,}814  & $-61.2$ & 12.3 & 4{,}769  & $-48.6$ \\
    \rowcolor[RGB]{242, 250, 246} ACTS$_{\pi_{\text{SFT}}}$   & 81.6 & 2{,}646  & $-51.3$ & 35.3 & 10{,}977 & $-34.6$ & 63.1 & 5{,}362  & $-50.5$ & 44.9 & 4{,}561  & $-63.3$ & 30.1 & 3{,}758  & $-59.5$ \\
    \rowcolor[RGB]{228, 247, 237} ACTS   & 82.8 & 2{,}532  & $-53.3$ & 36.7 & 10{,}417 & $-37.9$ & 65.1 & 5{,}143  & $-52.5$ & 45.9 & 4{,}460  & $-64.1$ & 30.1 & 3{,}766  & $-59.4$ \\
    \midrule
    \multicolumn{16}{l}{\textit{DeepSeek-R1-Distill-Qwen-7B}} \\
    \midrule
    Vanilla       & 92.6 & 4{,}339  & -     & 51.3 & 13{,}031 & -     & 83.1 & 7{,}863  & -     & 59.6 & 8{,}532  & -     & 38.9 & 8{,}422  & -     \\
    NoThink    & 78.8 & 730      & $-83.2$ & 25.3 & 3{,}944  & $-69.7$ & 54.5 & 1{,}600  & $-79.6$ & 44.7 & 1{,}645  & $-80.7$ & 35.4 & 700      & $-91.7$ \\
    CoD           & 78.6 & 1{,}650  & $-62.0$ & 46.0 & 11{,}427 & $-12.3$ & 67.0 & 4{,}431  & $-43.6$ & 45.8 & 5{,}716  & $-33.0$ & 36.7 & 5{,}402  & $-35.9$ \\
    BudgetGuidance & 82.8 & 2{,}294  & $-47.1$ & 50.7 & 11{,}694 & $-10.3$ & 78.8 & 6{,}600  & $-16.1$ & 56.7 & 6{,}890  & $-19.2$ & 36.5 & 4{,}565  & $-45.8$ \\
    DEER          & 77.8 & 2{,}029  & $-53.2$ & 50.0 & 9{,}411  & $-27.8$ & 71.3 & 4{,}891  & $-37.8$ & 51.0 & 4{,}994  & $-41.5$ & 34.9 & 3{,}825  & $-54.6$ \\
    \rowcolor[RGB]{242, 250, 246} ACTS$_{\pi_{\text{SFT}}}$   & 82.2 & 1{,}935  & $-55.4$ & 56.7 & 8{,}693  & $-33.3$ & 77.4 & 4{,}937  & $-37.2$ & 56.4 & 5{,}206  & $-39.0$ & 44.4 & 4{,}656  & $-44.7$ \\
    \rowcolor[RGB]{228, 247, 237} ACTS   & 85.2 & 1{,}866  & $-57.0$ & 60.0 & 8{,}200  & $-37.1$ & 79.8 & 4{,}486  & $-42.9$ & 58.1 & 4{,}181  & $-51.0$ & 46.8 & 4{,}404  & $-47.7$ \\
    \midrule
    \multicolumn{16}{l}{\textit{Qwen3-8B}} \\
    \midrule
    Vanilla       & 97.2 & 5{,}474  & -     & 76.0 & 14{,}880 & -     & 89.2 & 10{,}583 & -     & 67.7 & 11{,}368 & -     & 58.8 & 9{,}654  & -     \\
    NoThink    & 86.0 & 1{,}112  & $-79.7$ & 30.0 & 6{,}475  & $-56.5$ & 65.3 & 2{,}939  & $-72.2$ & 55.0 & 2{,}837  & $-75.0$ & 48.1 & 1{,}628  & $-83.1$ \\
    CoD           & 62.6 & 3{,}541  & $-35.3$ & 76.7 & 12{,}693 & $-14.7$ & 82.2 & 8{,}422  & $-20.4$ & 66.1 & 8{,}684  & $-23.6$ & 58.3 & 5{,}789  & $-40.0$ \\
    BudgetGuidance & 88.0 & 1{,}865  & $-65.9$ & 44.0 & 8{,}395  & $-43.6$ & 71.3 & 4{,}555  & $-57.0$ & 59.6 & 4{,}033  & $-64.5$ & 47.8 & 2{,}308  & $-76.1$ \\
    DEER          & 95.6 & 4{,}941  & $-9.7$  & 70.0 & 11{,}873 & $-20.2$ & 86.8 & 8{,}766  & $-17.2$ & 71.4 & 9{,}408  & $-17.2$ & 57.1 & 7{,}671  & $-20.5$ \\
    \rowcolor[RGB]{242, 250, 246} ACTS$_{\pi_{\text{SFT}}}$   & 94.0 & 3{,}573  & $-34.7$ & 74.0 & 12{,}727 & $-14.5$ & 87.2 & 8{,}371  & $-20.9$ & 67.4 & 8{,}799  & $-22.6$ & 58.4 & 7{,}761  & $-19.6$ \\
    \rowcolor[RGB]{228, 247, 237} ACTS   & 95.2 & 3{,}448  & $-37.0$ & 73.3 & 11{,}198 & $-24.7$ & 86.7 & 8{,}492  & $-19.8$ & 66.5 & 8{,}322  & $-26.8$ & 58.6 & 6{,}615  & $-31.5$ \\
    \bottomrule
    \end{tabular}%
    }
    \caption{Main results of accuracy (Acc), total token count (\#Tok), and token saving over Vanilla ($\Delta\%$) across five benchmarks and three reasoners.}
    \label{tab:main-results}
\end{table*}

\subsection{Baselines and Variants}

We compare ACTS against the following baselines spanning efficient reasoning, early exiting, and budget-aware reasoning regimes. \textbf{Vanilla} runs the frozen reasoner directly, producing a complete chain of thought followed by the final answer. \textbf{NoThink}~\cite{ma2025reasoning} prefills a closed empty thinking block at the start of the assistant turn, forcing the reasoner to skip the thinking phase and generate the answer directly. \textbf{CoD}~\cite{xu2025chain} prompts the reasoner to produce minimalistic intermediate reasoning steps, each limited to a few words. 
\textbf{DEER}~\cite{yang2026dynamic} dynamically truncates the chain of thought once the reasoner shows high confidence in a trial answer at reasoning-transition tokens. 
\textbf{BudgetGuidance}~\cite{li2025steering} steers token-level generation toward the thinking budget via an auxiliary lightweight predictor.
We reproduce all baselines using their released code under our experimental setup.
We evaluate a variant of our method \textbf{ACTS}$_{\pi_{\text{SFT}}}$, which uses our controller agents obtained from behavior initialization before reinforcement learning. 

\subsection{Implementation Details}

We instantiate the controller agent on Qwen3-4B-Instruct-2507~\cite{yang2025qwen3}. 
We train the controller via supervised finetuning on the constructed synthetic steering trajectories with learning rate $1\mathrm{e}{-5}$, global batch size $64$.
We further refine the controller via GRPO with learning rate $1\mathrm{e}{-6}$, group size $G{=}8$, rollout batch size $32$, train batch size $64$, penalty coefficient $\alpha{=}0.5$. 
During RL training, we pair the controller with DeepSeek-R1-Distill-Qwen-7B~\cite{guo2025deepseek} as the frozen reasoner. The controller is sampled with temperature $1.0$ and top-$p$ $0.9$, while the reasoner uses greedy decoding with temperature $0.0$ so that reward variance is attributable to controller decisions.
During evaluation, we set the controller temperature $0.7$ and top-$p$ $0.8$, and reasoner temperature $0.6$, top-$p$ $0.95$, and max new tokens $32{,}768$ following \citet{yang2026dynamic} for all compared baselines.
Controller SFT training uses OpenRLHF \cite{hu2024openrlhf} and RL training is implemented with SLIME \cite{slime_github}.
At inference, the controller and reasoner are served as separate SGLang~\cite{zheng2024sglang} servers communicating over asynchronous HTTP as detailed in Section~\ref{sec:inference}. 
The system prompts used by the controller and the reasoner are provided in Appendix~\ref{app:prompts}.
All experiments run on $8\times$A100 80GB GPUs.

\section{Results}

\begin{figure*}[t]
      \centering
      \includegraphics[width=\linewidth]{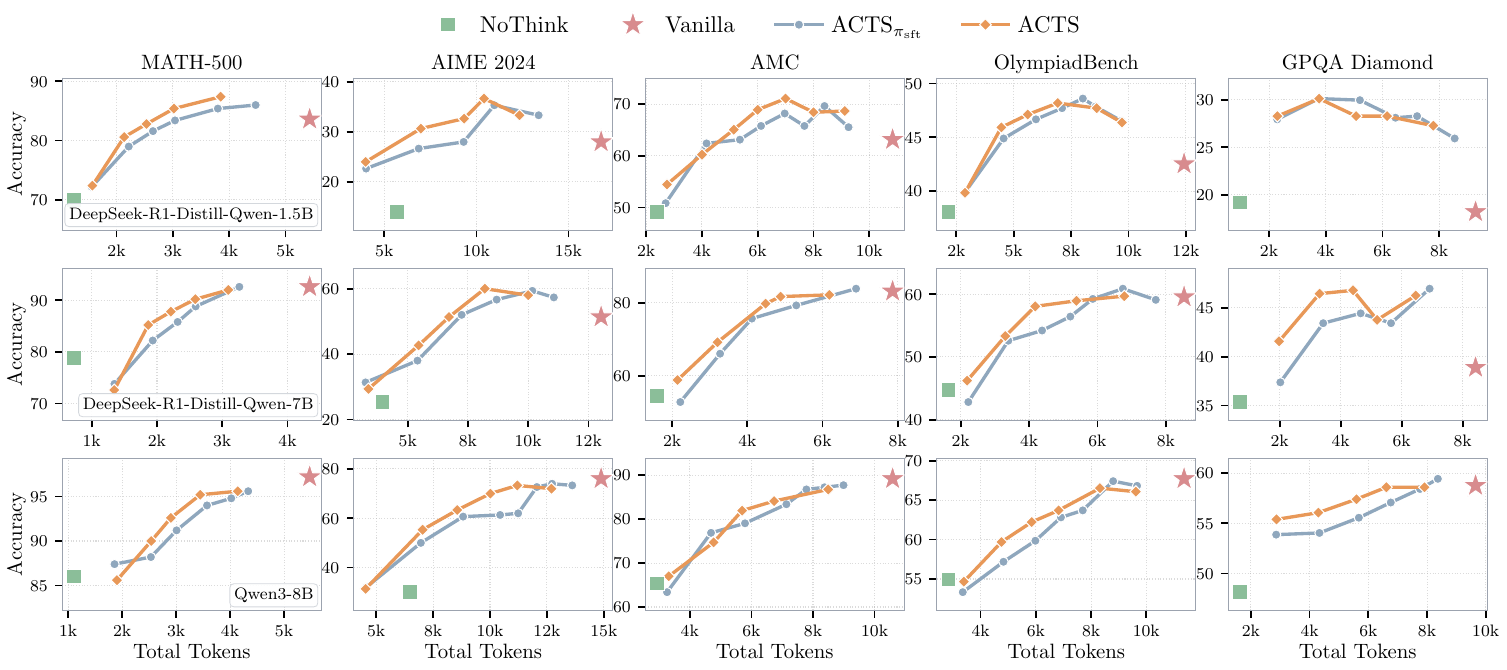}
      \caption{Accuracy vs. Total Tokens across three reasoners (rows) and five benchmarks (columns) under ACTS budget sweeps from tight to
  near-Vanilla scale.}
      \label{fig:pareto_curves}
\end{figure*}

\subsection{ACTS achieves competitive accuracy with substantial token savings}

We report in Table~\ref{tab:main-results} the average accuracy, total token count, and relative savings of ACTS against baselines across three reasoners and five benchmarks, with ACTS budgets set to scales comparable to baselines for fair comparison. 
From the results, we observe that ACTS provides favorable accuracy--token trade-offs across the tested settings. It improves over Vanilla in several harder settings for the DeepSeek reasoners, while on Qwen3-8B it trails Vanilla by only $0.2$--$2.7$ accuracy points with $20$--$37\%$ fewer tokens. ACTS also improves over the competitive budget-aware and efficient reasoning baselines BudgetGuidance and DEER on the majority of cells.
In contrast, NoThink and CoD, which directly suppress or compress the thinking phase, lose substantial accuracy on the harder benchmarks such as AIME, OlympiadBench, and GPQA, suggesting that uniformly suppressing the thinking phase trades away accuracy too aggressively on these tasks. In addition, ACTS$_{\pi_\text{sft}}$ with our SFT controller agent already approaches Vanilla accuracy at large savings, and ACTS with RL finetuning further boosts the controller agent performance.

\subsection{ACTS enables controllable accuracy vs.\ efficiency trade-offs}

As shown in Figure~\ref{fig:pareto_curves}, we sweep the ACTS thinking budget from tight to near-Vanilla scale and plot accuracy against total tokens for all three reasoners and five benchmarks. The curves are smooth and largely monotone, and on nearly all cells the ACTS curve lies above the line connecting the Vanilla and NoThink endpoints, indicating a strictly better accuracy-tokens trade-off than interpolating between the two extremes at any mixing ratio. With DeepSeek-R1-Distill-1.5B as the reasoner, ACTS even surpasses Vanilla on every harder benchmark, such as AIME, OlympiadBench, and GPQA, at substantially lower token cost, suggesting that structured steering can elevate weak reasoners above their unguided performance. 
The DeepSeek-R1-Distill-1.5B curves on AIME and GPQA also show a mild accuracy reversal at very large budgets. After analyzing the logged steering trajectories, we attribute this to the controller invoking an increasing number of \textsc{explore} actions (e.g., from $25\%$ to $42\%$ of all actions on GPQA as the budget triples) that the weak reasoner cannot sustain coherently, indicating that the budget signal itself plays a useful regularizing role for limited-capacity reasoners.

Additional ablations in Appendix~\ref{app:ablations} further examine the contributions of multi-budget augmentation, reward shaping, the strategy and steering phrase, and the remaining-budget signal.

\subsection{Our trained controller agent generalizes to different reasoners and tasks}

Our controller agent is trained only on synthetic steering trajectories derived from DeepSeek-R1 math reasoning, yet it transfers without retraining to a different reasoner, e.g., Qwen3-8B, and to out-of-domain science QA task as shown in Table \ref{tab:main-results} and Figure \ref{fig:pareto_curves}. Notably, ACTS achieves its largest gains on GPQA Diamond with the DeepSeek reasoners, improving over Vanilla by up to $+11.9$pp at substantially lower token cost despite never having seen science problems during training. By comparing Vanilla and controller-steered reasoning traces on GPQA, we attribute this transfer to a domain-agnostic failure mode of the unguided reasoners: their wrong attempts tend to run substantially longer than their correct ones, indicating that the reasoners frequently spiral into long, confused traces. Our controller agent's structured steering, drawing on the domain-agnostic strategy set $\mathcal{U}$ defined in Section~\ref{sec:mdp}, directly counteracts this overthinking by guiding the reasoner through discrete reasoning phases regardless of task content.
We further evaluate the same controller without retraining on
Llama-3.1-Nemotron-Nano-8B-v1. ACTS remains within $2.5$ accuracy points
of Vanilla across AIME24, AMC, and OlympiadBench while reducing token usage
by $18$--$20\%$. The full results are provided in Appendix~\ref{app:additional-results}.

\begin{figure}[t]
      \centering
      \includegraphics[width=\linewidth]{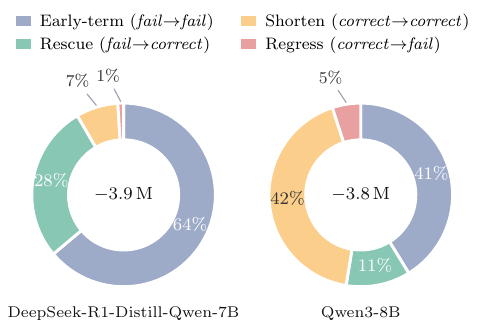}
      \caption{Decomposition of ACTS token savings.}
      \label{fig:token_savings}
\end{figure}

\subsection{Agentic reasoning steering offers interpretable token savings}

Beyond aggregate efficiency, ACTS's savings admit an interpretable per-trial decomposition. We classify every problem trial by the joint correctness of Vanilla and ACTS into four outcome categories: (i) Early-term, i.e., both wrong, (ii) Rescue, i.e., Vanilla wrong while ACTS correct, (iii) Shorten, i.e., both correct, and (iv) Regress, i.e., Vanilla correct while ACTS wrong. We choose ACTS runs at the smallest budget that matches Vanilla accuracy. 
As shown in Figure~\ref{fig:token_savings}, the savings concentrate in mechanisms that reflect genuine reasoning steering rather than indiscriminate truncation. On DeepSeek-R1-7B, $28\%$ of the savings come from Rescue, where the controller turns a wrong answer correct while still using fewer tokens; this is what enables ACTS to match or even outperform Vanilla accuracy under a tighter budget. On the stronger Qwen3-8B, where rescues are rare, $42\%$ of the savings instead come from Shorten, where the controller trims post-solution verification detours and strategy oscillations that the reasoner would otherwise explore. For both reasoners, Regress contributes less than $5\%$ of the savings, indicating that the controller learns effective steering behaviors rather than uniformly clipping reasoning length.

We illustrate both mechanisms with examples detailed in Appendix \ref{app:case-studies}. For Rescue, Vanilla arrives at the correct candidate but makes an error during self-verification, abandons it, and spirals into a long confused trace before settling on a wrong answer; ACTS instead invokes a sequence of \textsc{check} actions to systematically confirm the candidate and concludes correctly with a small fraction of the tokens. For Shorten, both methods reach the correct answer, but Vanilla then detours into an unnecessary post-solution verification and continues generating redundant tokens, while ACTS follows a clean \textsc{plan}-\textsc{execute}-\textsc{check}-\textsc{conclude} path and stops promptly with large token savings.

\subsection{Async controller-reasoner inference maintains competitive throughput}\label{sec:latency-analysis}

\begin{figure}[t]
      \centering
      \includegraphics[width=\linewidth]{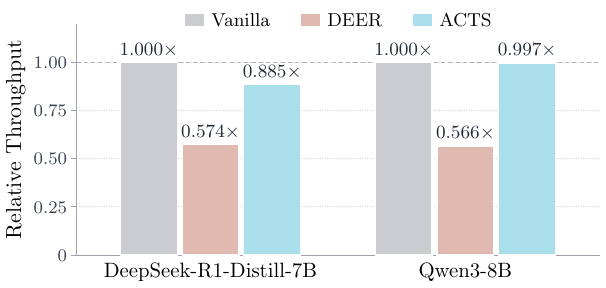}
      \caption{Inference throughput (\#Tok$/$s) comparisons.}
      \label{fig:latency_comparison}
\end{figure}

One practical deployment concern of our ACTS framework is inference latency, since the controller-reasoner architecture introduces additional controller calls on top of reasoner generation. To measure how our asynchronous two-server pipeline in Section~\ref{sec:inference} addresses this concern, we benchmark end-to-end token throughput (\#Tokens per second) on MATH-500 against the Vanilla and DEER baselines. All three compared methods are implemented with SGLang inference engine for a fair comparison, and share the same $8$-GPU compute budget: Vanilla and DEER use all $8$ GPUs for the reasoner, while ACTS splits them into $4{+}4$ between the controller and reasoner. As shown in Figure~\ref{fig:latency_comparison}, ACTS matches Vanilla's throughput within $1\%$ on Qwen3-8B and within $11\%$ on DeepSeek-R1-Distill-7B, indicating that asynchronous serving effectively amortizes the controller-reasoner round trips. In contrast, DEER reaches only about half of Vanilla's throughput on both reasoners, as its iterative probe-and-resume cycles add overhead that batched serving cannot hide.

Using the standard generation FLOPs approximation $2\times\text{\#parameters}\times\text{\#generated tokens}$ and accounting for both controller and reasoner generation, ACTS uses approximately $74\%$ of Vanilla's estimated FLOPs with the $7$B and $8$B reasoners, but approximately $99\%$ with the $1.5$B reasoner, where the $4$B controller largely offsets the reasoner-token savings.

\section{Conclusion}

In this paper, we present Agentic Chain-of-Thought Steering (ACTS), a framework that formulates reasoning steering as a controller-reasoner Markov decision process under a thinking-token budget. At each step, the controller observes the steering history and remaining budget, and emits a reasoning strategy and a steering phrase that initiates the next reasoner step. We train it in two stages: behavior initialization from our constructed synthetic steering trajectories with multi-budget augmentation, followed by reinforcement learning with budget-conditioned reward shaping that asymmetrically penalizes overthinking and premature termination. Experiments show that ACTS achieves competitive accuracy with substantial token savings, enables controllable accuracy--token trade-offs, and transfers across the tested reasoners and task domains. Our results on strategy-level steering offer a complementary control surface for efficient reasoning beyond global length control.

\section*{Limitations}

While ACTS demonstrates effective and controllable reasoning steering, our study has several limitations. First, our experiments cover three reasoners ranging up to $8$B parameters from two model families. Whether the same controller can steer substantially larger reasoners, such as $70$B-scale open-weight models or frontier proprietary reasoners, is beyond our compute budget for evaluation. While our analysis indicates that the controller exploits relatively domain- and model-agnostic failure modes such as overthinking, we leave direct validation at larger scales for future investigation. Second, ACTS assumes that the thinking-token budget is supplied at inference time by the user or the service provider. This matches deployment settings where budgets are externally constrained by latency targets or cost ceilings, while a fully autonomous variant would infer a problem-difficulty-appropriate budget on its own. We leave the design of such an adaptive budget mechanism, e.g., via a difficulty estimator or a learned early-confidence signal, for future work.

\section*{Acknowledgments}
This work is supported by the Intuit University Collaboration Program.

\bibliography{anthology-1, custom}

\appendix

\section*{Appendix}

\section{Strategies and Steering Phrases}
\label{app:strategies}

Table~\ref{tab:strategies} summarizes the seven reasoning strategies that constitute the controller's strategy set $\mathcal{U}$, alongside example steering phrases observed in our synthetic SFT corpus. Each strategy $u_t \in \mathcal{U}$ specifies a high-level functional role. 
The steering phrase $p_t$ provides the opening tokens of the next reasoner step, linguistically enacting the chosen strategy while preserving continuity with the reasoner's native generation style. 
The example phrases are illustrative and not exhaustive: in practice the controller emits free-form phrases conditioned on the running trace, and the same strategy can be entered through many different surface forms.

\begin{table*}[!t]
\centering
\renewcommand{\arraystretch}{1.35}
\small
\begin{tabular}{l p{5.2cm} p{7.5cm}}
\toprule
\textbf{Strategy} & \textbf{Functional role} & \textbf{Example steering phrases} \\
\midrule
\textsc{understand} & Restate or clarify the problem. & \emph{``Okay, so I have this problem here:''}, \emph{``Wait, the problem states:''}, \emph{``Let me rephrase the problem.''} \\
\textsc{plan}       & Decide overall strategy or split into subproblems. & \emph{``Alright, let's see.''}, \emph{``First, I need to find\ldots''}, \emph{``Let me set up\ldots''} \\
\textsc{execute}    & Carry out algebraic or logical derivations. & \emph{``Next,\ldots''}, \emph{``Moving on to\ldots''}, \emph{``So for\ldots''} \\
\textsc{explore}    & Try an alternative approach or angle. & \emph{``Alternatively, let's approach it\ldots''}, \emph{``Maybe there's another way.''}, \emph{``But perhaps\ldots''} \\
\textsc{check}      & Verify intermediate steps or correct mistakes. & \emph{``Wait, let me verify\ldots''}, \emph{``Let me double-check\ldots''}, \emph{``Wait, let's recheck.''} \\
\textsc{summarize}  & Summarize partial results so far. & \emph{``Therefore, the sum\ldots''}, \emph{``So we have\ldots''}, \emph{``In this case,\ldots''} \\
\textsc{conclude}   & State the final answer and terminate the thinking trace. & \emph{``\texttt{**Final Answer**}\ldots''} \\
\bottomrule
\end{tabular}
\caption{The controller's strategy set $\mathcal{U}$, with each strategy's functional role and representative steering phrases from our constructed synthetic steering trajectory corpus.}
\label{tab:strategies}
\end{table*}

\begin{table*}[!t]
\centering
\resizebox{\textwidth}{!}{%
\begin{tabular}{lccccc}
\toprule
\textbf{Method} &
\textbf{2{,}499} &
\textbf{4{,}998} &
\textbf{7{,}498} &
\textbf{9{,}997} &
\textbf{12{,}497} \\
\midrule
ACTS
& 29.3 (3{,}366)
& 42.7 (5{,}451)
& 51.3 (6{,}711)
& 60.0 (8{,}200)
& 58.0 (10{,}012) \\
w/o multi-budget augmentation
& 44.7 (11{,}411)
& 46.0 (12{,}762)
& 53.3 (11{,}801)
& 55.3 (12{,}294)
& 54.0 (13{,}014) \\
w/o budget-underuse reward shaping
& 31.3 (3{,}033)
& 38.0 (4{,}549)
& 42.7 (5{,}463)
& 46.7 (5{,}550)
& 42.0 (5{,}814) \\
\bottomrule
\end{tabular}%
}
\caption{Training-design ablations across different thinking-token budgets. Each entry reports Accuracy (\#Tokens).}
\label{tab:training-ablation}
\end{table*}

\begin{table*}[!t]
\centering
\resizebox{\textwidth}{!}{%
\begin{tabular}{lccccc}
\toprule
\textbf{Method} &
\textbf{2{,}499} &
\textbf{4{,}998} &
\textbf{7{,}498} &
\textbf{9{,}997} &
\textbf{12{,}497} \\
\midrule
ACTS$_{\pi_{\mathrm{SFT}}}$
& 31.3 (3{,}248)
& 38.0 (5{,}403)
& 52.0 (7{,}238)
& 56.7 (8{,}693)
& 59.3 (10{,}174) \\
w/ learned strategy + templated phrase
& 28.0 (3{,}285)
& 38.0 (5{,}409)
& 43.3 (7{,}530)
& 49.3 (8{,}597)
& 50.0 (9{,}984) \\
w/ learned phrase only
& 28.0 (3{,}243)
& 38.0 (5{,}180)
& 44.0 (6{,}826)
& 51.3 (8{,}028)
& 51.3 (9{,}496) \\
w/ random strategy + learned phrase
& 40.7 (12{,}552)
& 39.3 (12{,}497)
& 43.3 (12{,}785)
& 41.3 (11{,}448)
& 44.0 (12{,}317) \\
\bottomrule
\end{tabular}%
}
\caption{Strategy and steering-phrase ablations across different thinking-token budgets. Each entry reports Accuracy (\#Tokens).}
\label{tab:action-ablation}
\end{table*}

\section{Steering Trajectory Construction}
\label{app:sft}

We construct synthetic steering trajectories from OpenR1-Math~\cite{openr1} with reasoning traces produced by DeepSeek-R1~\cite{guo2025deepseek}. To cover a broad range of trace lengths, we length-balance the source by sampling $500$ traces from each of $15$ non-overlapping 512-token bands over $[512,8192]$ trace tokens, yielding $7{,}500$ source traces. Each trace is segmented into reasoning steps at paragraph boundaries that end in sentence-final punctuation (\texttt{.\textbackslash n\textbackslash n} or \texttt{?\textbackslash n\textbackslash n}).

\paragraph{Action annotation and extraction.} For each step, we use Qwen3-32B as an LLM annotator that jointly (i) classifies the step into one of seven strategies $\{$\textsc{understand}, \textsc{plan}, \textsc{execute}, \textsc{explore}, \textsc{check}, \textsc{summarize}, \textsc{conclude}$\}$ and (ii) extracts the step's opening as steering phrase. The annotator prompt is provided in Figure~\ref{fig:extraction-prompt}. We use temperature $0.7$, top-$p$ $0.8$, and top-$k$ $20$ and the annotator returns a JSON object which we parse directly into the strategy $u_t$ and steering phrase $p_t$.

\paragraph{Trajectory assembling.}
Each annotated trace is assembled into a multi-turn dialog matching the controller's chat format at inference time, with the controller system prompt (Figure~\ref{fig:controller-system-prompt}) as the system message and the annotator's \emph{Action / Starter} pair as the assistant target at each turn. Figure~\ref{fig:trajectory-example} shows a worked example from our SFT corpus. During SFT, the loss is computed only on assistant tokens.

\paragraph{Trajectory filtering.}
After assembling steering trajectories, we apply two filters to suppress degenerate looping behavior. First, consecutive same-strategy runs in each trajectory are capped at $5$ to discourage repetitive sequences. Second, trajectories that contain $5$ or more identical opening steering phrases or exhibit more than $5$ alternating-strategy cycles are dropped from the corpus.
Starting from $74{,}990$ OpenR1-Math traces, length-balanced sampling yields
$7{,}500$ source traces. Annotation and filtering retain $5{,}768$ traces
($23.1\%$ filtered), and multi-budget augmentation produces $17{,}277$
steering trajectories. We use sampling seed $42$ for Qwen3-32B annotation.
Re-annotating $500$ steps with a fresh seed of $1337$ yields Cohen's
$\kappa=0.899$. We release the resulting synthetic steering trajectories
and annotation outputs with our code.

\section{Ablation Studies}
\label{app:ablations}

We conduct ablations on AIME24 with DeepSeek-R1-Distill-Qwen-7B as the frozen reasoner to examine the contributions of our training design, steering action, and remaining-budget signal.

\paragraph{Training Design.}
Table~\ref{tab:training-ablation} compares ACTS with variants removing multi-budget augmentation or the budget-underuse term in reward shaping. Without multi-budget augmentation, the controller consistently exceeds tight budget targets and fails to adapt its token usage across the budget sweep. Without the budget-underuse penalty, the controller instead saturates at around $5.5$K tokens even as more budget becomes available, leading to lower accuracy at larger budgets. These results show that the two designs address complementary failure modes in budget-aware control.

\paragraph{Strategy and Steering Phrase.}
We further isolate the two components of the controller action using the SFT-initialized controller. The results are shown in Table \ref{tab:action-ablation}. Replacing the learned steering phrase with the most frequent phrase for each strategy degrades accuracy, as does removing the explicit strategy label and training the controller to emit only the steering phrase. Forcing a random strategy at inference further disrupts budget compliance. Together, these results indicate that both the explicit reasoning strategy and the learned free-form steering phrase contribute to ACTS.

\paragraph{Remaining-Budget Signal.}
Finally, we test whether the controller actually uses the explicit \texttt{Budget Remaining: X\%} signal. We either replace it with a random percentage independently at each controller turn (\textsc{Budget-Random}) or remove the budget line entirely (\textsc{Budget-Removed}). Across the same budget sweep, \textsc{Budget-Random} plateaus at $34.7\%$ accuracy with approximately $5$K tokens, while \textsc{Budget-Removed} plateaus at $53.9\%$ accuracy with approximately $11$K tokens. In both cases, actual token usage becomes largely decoupled from the requested budget, confirming that the controller uses the remaining-budget signal for budget-conditioned reasoning control.

\section{Additional Results}
\label{app:additional-results}

\paragraph{ACTS Budgets in Main Results.}
Table~\ref{tab:main_budget} reports the thinking-token budget used for each ACTS result in Table~\ref{tab:main-results}. These operating points are selected from the budget sweeps in Figure~\ref{fig:pareto_curves}, whose budgets range from $0.2\times$ to $1.6\times$ the corresponding Vanilla mean thinking-token length.

\begin{table*}[!t]
\centering
\resizebox{\textwidth}{!}{%
\begin{tabular}{lccccc}
\toprule
\textbf{Reasoner} & \textbf{MATH-500} & \textbf{AIME24} & \textbf{AMC} & \textbf{OlympiadBench} & \textbf{GPQA Diamond} \\
\midrule
DeepSeek-R1-Distill-Qwen-1.5B & 3{,}021 & 16{,}182 & 6{,}224 & 4{,}786 & 3{,}554 \\
DeepSeek-R1-Distill-Qwen-7B   & 1{,}580 & 9{,}997  & 5{,}889 & 4{,}815 & 4{,}770 \\
Qwen3-8B                      & 4{,}836 & 13{,}755 & 11{,}671 & 12{,}415 & 7{,}032 \\
\bottomrule
\end{tabular}%
}
\caption{ACTS thinking-token budgets used for the results in Table~\ref{tab:main-results}.}
\label{tab:main_budget}
\end{table*}

\paragraph{Additional Cross-Model Generalization.}
To further evaluate transfer beyond the DeepSeek and Qwen reasoners, we pair the same trained controller, without additional training, with Llama-3.1-Nemotron-Nano-8B-v1 \cite{bercovich2025llamanemotronefficientreasoningmodels}. As shown in Table~\ref{tab:nemotron}, ACTS remains within $2.5$ accuracy points of Vanilla across AIME24, AMC, and OlympiadBench while reducing token usage by $18$--$20\%$.

\begin{table}[!t]
\centering
\resizebox{\columnwidth}{!}{%
\begin{tabular}{lcccc}
\toprule
\textbf{Benchmark} & \textbf{Vanilla} & \textbf{ACTS} & $\Delta$\textbf{Acc} & \textbf{Savings} \\
\midrule
AIME24
& 62.7 (11{,}248)
& 63.3 (9{,}232)
& $+0.6$
& $18\%$ \\
AMC
& 86.0 (7{,}316)
& 84.6 (5{,}856)
& $-1.4$
& $20\%$ \\
OlympiadBench
& 64.9 (7{,}517)
& 62.4 (6{,}152)
& $-2.5$
& $18\%$ \\
\bottomrule
\end{tabular}%
}
\caption{Additional cross-model generalization results. Accuracy and token usage are shown as Acc (\#Tokens).}
\label{tab:nemotron}
\end{table}

\section{Prompts for Controller \& Reasoner}
\label{app:prompts}

\paragraph{Controller System Prompt.}
The system prompt for the controller agent is provided in Figure~\ref{fig:controller-system-prompt}, which is used in SFT, RL training, and evaluation.

\paragraph{Controller User Messages.}
On the first turn of the controller agent, the user message provides the question together with the full budget.
On subsequent turns, the user message reports the reasoner's latest step and the current remaining-budget fraction. 
The two templates are shown in Figure~\ref{fig:controller-user-messages}. 
The controller's assistant response follows the output format specified in the system prompt: the strategy $u_t$ is parsed from the \texttt{Action:} line and the steering phrase $p_t$ from the \texttt{Starter:} line, jointly forming the steering action $a_t = (u_t, p_t)$.

\paragraph{Reasoner Prompt.}
The reasoner uses no custom system prompt by default. For each step $t$, we apply the model's native chat template to the question, then append the cumulative reasoning trace $z_{<t}$ from prior steps inside the \texttt{<think>} block, followed by the controller's steering phrase $p_t$ that opens step $t$. Figure~\ref{fig:reasoner-prompt} shows the full construction. This preserves the reasoner's native generation continuity while letting the controller steer each subsequent step through a short appended steering phrase.

\balance
\section{Case Studies}
\label{app:case-studies}

We present two MATH-500 examples in Figure \ref{fig:case-rescue} and Figure \ref{fig:case-shorten} illustrating the Rescue and Shorten mechanisms of ACTS. Both compare the unguided Vanilla reasoner with our controller agent paired with DeepSeek-R1-Distill-Qwen-7B. Verbose Vanilla portions are abbreviated as \textit{[\ldots\ XX tokens omitted: brief description \ldots]}; ACTS steering phrases generated by the controller agent are marked in orange.

\begin{figure*}[!t]
    \definecolor{promptback}{RGB}{252, 252, 252}
    \definecolor{promptframe}{RGB}{138, 147, 156}

\begin{tcblisting}{%
  breakable,
  enhanced jigsaw,
  colback=promptback,
  colframe=promptframe,
  coltitle=white,
  fonttitle=\bfseries,
  arc=2pt,
  boxrule=0.5pt,
  leftrule=0.5pt,
  rightrule=0.5pt,
  toprule=0.5pt,
  bottomrule=0.5pt,
  title={Annotator Prompt},
  listing only
}
Classify this reasoning step by its opening intent. Choose exactly one action.

- "understand": Re-reads or clarifies the problem, e.g., So, the problem..., The problem says....
- "plan": Outlines strategy, defines variables, or enumerates cases, e.g., First, let me..., Let me consider...
- "execute": Performs computation, algebra, or derivation, e.g., Compute the value..., From the equation...
- "explore": Abandons the current path for a new approach, e.g., Alternatively, maybe it's..., Alternatively, perhaps the...
- "check": Pauses and then questions or verifies a specific prior result, e.g., Wait, let me..., Wait, hold on,...
- "summarize": Restates what has been established so far, e.g., Therefore, the number..., Thus, the total...
- "conclude": States the final answer and terminates the trace, e.g., **Final Answer**...

Then extract the short opening phrase of this reasoning step: the question-agnostic discourse marker or setup phrase at the start of the step; typically 1-8 words; terminating at a natural linguistic boundary; return it verbatim.

Reasoning step:
---
{step}
---

Respond with a JSON object:
{"action": "action_category", "starter": "<extracted opening phrase>"}
\end{tcblisting}
\caption{Prompt used by the annotator to jointly classify each reasoning step into one of seven strategies and extract its opening starter phrase.}
\label{fig:extraction-prompt}
\end{figure*}

\begin{figure*}[!t]
    \definecolor{promptback}{RGB}{252, 252, 252}
    \definecolor{promptframe}{RGB}{138, 147, 156}

\begin{tcblisting}{%
  breakable,
  enhanced jigsaw,
  colback=promptback,
  colframe=promptframe,
  coltitle=white,
  fonttitle=\bfseries,
  arc=2pt,
  boxrule=0.5pt,
  leftrule=0.5pt,
  rightrule=0.5pt,
  toprule=0.5pt,
  bottomrule=0.5pt,
  title={Controller System Prompt},
  listing only
}
You are a reasoning control agent that guides a step-by-step
mathematical reasoner.

Each turn you receive the reasoner's latest step and the
remaining token budget (as a percentage). On the first turn
you receive the question instead. Considering the reasoner's
current progress and planning on the remaining budget, choose
an action and a starter phrase for the reasoner's next step.
Your goal is to guide the reasoner to make progress towards
the final correct answer while staying within the budget.

Actions:
- understand: Restate or clarify the problem.
- plan:       Decide strategy or split into subproblems.
- execute:    Carry out algebraic or logical steps.
- explore:    Try an alternative approach.
- check:      Verify steps or correct mistakes.
- summarize:  Summarize current results.
- conclude:   State the final answer.

Output format:
Action: <action>
Starter: <phrase that steers the reasoner toward the chosen action>
\end{tcblisting}

\caption{System prompt for the controller agent.}
\label{fig:controller-system-prompt}
\end{figure*}

\begin{figure*}[!t]
    \definecolor{promptback}{RGB}{252, 252, 252}
    \definecolor{promptframe}{RGB}{138, 147, 156}

\begin{tcblisting}{%
  breakable,
  enhanced jigsaw,
  colback=promptback,
  colframe=promptframe,
  coltitle=white,
  fonttitle=\bfseries,
  arc=2pt,
  boxrule=0.5pt,
  leftrule=0.5pt,
  rightrule=0.5pt,
  toprule=0.5pt,
  bottomrule=0.5pt,
  title={Controller User Message: First Turn},
  listing only
}
Question:
{question}

Budget Remaining: 100%
\end{tcblisting}

\vspace{4pt}

\begin{tcblisting}{%
  breakable,
  enhanced jigsaw,
  colback=promptback,
  colframe=promptframe,
  coltitle=white,
  fonttitle=\bfseries,
  arc=2pt,
  boxrule=0.5pt,
  leftrule=0.5pt,
  rightrule=0.5pt,
  toprule=0.5pt,
  bottomrule=0.5pt,
  title={Controller User Message: Subsequent Turns},
  listing only
}
Reasoner's Step:
{current_step}

Budget Remaining: {b_t}%
\end{tcblisting}

\caption{Controller user message templates. The first turn supplies the question with full budget; subsequent turns report the reasoner's latest step and the current remaining budget.}
\label{fig:controller-user-messages}
\end{figure*}

\begin{figure*}[!t]
    \definecolor{promptback}{RGB}{247, 247, 247}
    \definecolor{promptframe}{RGB}{138, 147, 156}

\begin{tcblisting}{%
  breakable,
  enhanced jigsaw,
  colback=promptback,
  colframe=promptframe,
  coltitle=white,
  fonttitle=\bfseries,
  arc=2pt,
  boxrule=0.5pt,
  leftrule=0.5pt,
  rightrule=0.5pt,
  toprule=0.5pt,
  bottomrule=0.5pt,
  title={Reasoner Prompt at Step $t$},
  listing only
}
{question}
<think>
{cumulative trace z_<t}{steering phrase p_t}
\end{tcblisting}

\caption{Reasoner prompt construction at step $t$. The reasoner inherits the model's native chat template; only the thinking trace and steering phrase generated by the controller agent are appended after \texttt{<think>}.}
\label{fig:reasoner-prompt}
\end{figure*}

\begin{figure*}[!t]
    \definecolor{promptback}{RGB}{252, 252, 252}
    \definecolor{promptframe}{RGB}{138, 147, 156}
    \definecolor{actsblue}{RGB}{30,80,180}
    \definecolor{actsgreen}{RGB}{40,130,80}
    \definecolor{actsgray}{RGB}{100,100,100}
    \definecolor{actsplum}{RGB}{140,60,140}
    
    \begin{tcolorbox}[
      enhanced jigsaw,
      colback=promptback,
      colframe=promptframe,
      coltitle=white,
      fonttitle=\bfseries,
      arc=2pt,
      boxrule=0.5pt,
      title={From Reasoning Trace to Steering Trajectory},
      overlay={
        \draw[gray!55, line width=0.4pt, dashed]
          ([yshift=-4pt]interior.north) -- ([yshift=4pt]interior.south);
      },
    ]
    \noindent
    \begin{minipage}[t]{0.475\textwidth}
    \footnotesize
    \setlength{\parskip}{3pt}
    \begin{center}{\textbf{\normalsize Annotated Reasoning Trace}}\end{center}
    
    \textbf{Question:} Which is greater, $\log_{2008}(2009)$ or $\log_{2009}(2010)$?\\[3pt]
    
    \textcolor{actsplum}{\textbf{[Step 1 $\,\vert\,$ understand]}}\\
    \underline{Okay, so I need to figure out whether} $\log_{2008}(2009)$ is greater than $\log_{2009}(2010)$, or vice versa. Both logs are of consecutive integers with shifted bases.\\[3pt]
    
    \textcolor{actsplum}{\textbf{[Step 2 $\,\vert\,$ plan]}}\\
    \underline{First,} the logarithm $\log_b(a)$ is increasing in $a$ when the base $b > 1$. Recall the change of base: $\log_b(a) = \ln(a)/\ln(b)$.\\[3pt]
    
    \textcolor{actsplum}{\textbf{[Step 3 $\,\vert\,$ explore]}}\\
    \underline{Alternatively, maybe I can use} the change of base formula to convert both logs to natural logs and compare via the ratio $A/B$.\\[3pt]
    
    \textcolor{actsplum}{\textbf{[Step 4 $\,\vert\,$ execute]}}\\
    $\underline{A/B} = \dfrac{\ln(2009)/\ln(2008)}{\ln(2010)/\ln(2009)} = \dfrac{[\ln(2009)]^2}{\ln(2008)\cdot\ln(2010)}$, so we need $[\ln(2009)]^2 > \ln(2008)\cdot\ln(2010)$.\\[3pt]
    
    \textcolor{actsplum}{\textbf{[Step 5 $\,\vert\,$ explore]}}\\
    \underline{Alternatively, consider} $f(x) = \log_x(x+1)$. If $f$ is decreasing, then the larger base yields the smaller value.\\[3pt]

    \textcolor{actsplum}{\textbf{[Step 6 $\,\vert\,$ execute]}}\\
    \underline{To check if} $f(x)$ is monotonic, compute its derivative with respect to $x$.\\[3pt]

    \begin{center}\textcolor{actsgray}{\textit{$\vdots$ \;\, (16 steps omitted: derivative of $f$, concavity of $\ln$, verifying with $x=2,3$) \;\, $\vdots$}}\end{center}
    
    
    \textcolor{actsplum}{\textbf{[Step 24 $\,\vert\,$ summarize]}}\\
    \underline{So,} after derivatives, concavity, and algebraic manipulation, the first logarithm is consistently greater than the second.\\[3pt]
    
    \textcolor{actsplum}{\textbf{[Step 25 $\,\vert\,$ conclude]}}\\
    \underline{\texttt{**Final Answer**}} $\log_{2008}(2009) > \log_{2009}(2010)$.
    \end{minipage}\hfill
    \begin{minipage}[t]{0.475\textwidth}
    \footnotesize
    \setlength{\parskip}{3pt}
    \begin{center}{\textbf{\normalsize Assembled Steering Trajectory}}\end{center}
    
    \textcolor{actsgray}{\textbf{[System]}} \emph{Controller agent system prompt} (Figure~\ref{fig:controller-system-prompt})\\[3pt]
    
    \textcolor{actsblue}{\textbf{[User]}} Question: Which is greater, $\log_{2008}(2009)$ or $\log_{2009}(2010)$?\\
    Budget Remaining: 100\%\\
    \textcolor{actsgreen}{\textbf{[Assistant]}} Action: \texttt{understand}\\
    Starter: \underline{Okay, so I need to figure out whether}\\[3pt]
    
    \textcolor{actsblue}{\textbf{[User]}} Reasoner's Step: Okay, so I need to figure out whether $\log_{2008}(2009)$\ldots\\
    Budget Remaining: 98\%\\
    \textcolor{actsgreen}{\textbf{[Assistant]}} Action: \texttt{plan}\\
    Starter: \underline{First,}\\[3pt]
    
    \textcolor{actsblue}{\textbf{[User]}} Reasoner's Step: First, the logarithm $\log_b(a)$ is increasing\ldots change of base.\\
    Budget Remaining: 93\%\\
    \textcolor{actsgreen}{\textbf{[Assistant]}} Action: \texttt{explore}\\
    Starter: \underline{Alternatively, maybe I can use}\\[3pt]
    
    \textcolor{actsblue}{\textbf{[User]}} Reasoner's Step: Alternatively, maybe I can use the change of base\ldots ratio $A/B$.\\
    Budget Remaining: 86\%\\
    \textcolor{actsgreen}{\textbf{[Assistant]}} Action: \texttt{execute}\\
    Starter: \underline{A/B}\\[3pt]
    
    \textcolor{actsblue}{\textbf{[User]}} Reasoner's Step: $A/B = [\ln(2009)]^2 / [\ln(2008)\cdot\ln(2010)]$\ldots\\
    Budget Remaining: 82\%\\
    \textcolor{actsgreen}{\textbf{[Assistant]}} Action: \texttt{explore}\\
    Starter: \underline{Alternatively, consider}\\[3pt]

    \textcolor{actsblue}{\textbf{[User]}} Reasoner's Step: Alternatively, consider $f(x) = \log_x(x+1)$\ldots\\
    Budget Remaining: 79\%\\
    \textcolor{actsgreen}{\textbf{[Assistant]}} Action: \texttt{execute}\\
    Starter: \underline{To check if}\\[3pt]

    \begin{center}\textcolor{actsgray}{\textit{$\vdots$ \;\, (16 controller--reasoner turns omitted) \;\, $\vdots$}}\end{center}
    
    \textcolor{actsblue}{\textbf{[User]}} Reasoner's Step: Expressed in terms of $d_1, d_2$\ldots $d_1 > d_2$ by concavity.\\
    Budget Remaining: 3\%\\
    \textcolor{actsgreen}{\textbf{[Assistant]}} Action: \texttt{summarize}\\
    Starter: \underline{So,}\\[3pt]
    
    \textcolor{actsblue}{\textbf{[User]}} Reasoner's Step: So, after going through different approaches\ldots first log is greater.\\
    Budget Remaining: 1\%\\
    \textcolor{actsgreen}{\textbf{[Assistant]}} Action: \texttt{conclude}\\
    Starter: \underline{\texttt{**Final Answer**}}
    \end{minipage}
    \end{tcolorbox}
    \caption{An example of assembled steering trajectory (right) constructed from an annotated reasoning trace (left). The annotated strategies are noted in purple and the extracted steering phrases are underlined.}
    \label{fig:trajectory-example}
\end{figure*}

\begin{figure*}[!t]
    \definecolor{promptback}{RGB}{252, 252, 252}
    \definecolor{promptframe}{RGB}{138, 147, 156}
    \definecolor{actsorange}{RGB}{210, 120, 30}
    \definecolor{actsred}{RGB}{170, 60, 60}
    \definecolor{actsgray}{RGB}{100, 100, 100}

    \begin{tcolorbox}[
      enhanced jigsaw,
      colback=promptback,
      colframe=promptframe,
      coltitle=white,
      fonttitle=\bfseries,
      arc=2pt,
      boxrule=0.5pt,
      title={Case 1: Rescue},
      overlay={
        \draw[gray!55, line width=0.4pt, dashed]
          ([yshift=-4pt]interior.north) -- ([yshift=4pt]interior.south);
      },
    ]
    \noindent
    {\small\textbf{Problem.} What is the smallest positive multiple of $450$ whose digits are all zeroes and ones?\\\textbf{Ground truth:} $11{,}111{,}111{,}100$}
    \vspace{2pt}

    \noindent
    \begin{minipage}[t]{0.475\textwidth}
    \scriptsize
    \setlength{\parskip}{3pt}
    \begin{center}{\small\textbf{Vanilla}}\end{center}

    \textcolor{actsgray}{\texttt{<think>}}

    Alright, so I need to find the smallest positive multiple of $450$ where all the digits are either $0$ or $1$. Hmm, okay. Let's break this down.

    First, I know that $450 = 45 \times 10$, so the number must be divisible by both $45$ and $10$. 
    
    Divisibility by $10$: the number must end with a $0$. Divisibility by $9$: the sum of digits must be a multiple of $9$. Since digits are $0$ or $1$, the digit sum is the count of $1$'s, so the count of $1$'s must be a multiple of $9$.

    So our number must end with $0$ and have a count of $1$'s that is a multiple of $9$. The smallest such number is $1111111110$.

    \textcolor{actsred}{Wait, but hold on---is that the smallest?} Let me check. $1111111110$ ends with a single $0$, but $450$ requires divisibility by $25$, so the last two digits must be $00$. So $1111111110$ is invalid.

    Updating: the number must end with two $0$'s. So the smallest candidate is $11111111100$ (nine $1$'s followed by $00$). Let me verify: digit sum is $9$ \checkmark; ends in $00$ \checkmark. So $11111111100$ is indeed a multiple of $450$.

    \textcolor{actsred}{But is this the smallest? Maybe there's a number with fewer digits.} In $10$ digits ending with $00$, the maximum digit sum is $8$ (eight $1$'s and two $0$'s), which is not a multiple of $9$. So no $10$-digit candidate works.

    \textcolor{actsred}{Let me also try a brute-force enumeration of $450 \cdot k$ for small $k$\ldots} $k=1$: $450$. $k=2$: $900$. $k=3$: $1350$. $k=4$: $1800$. $k=5$: $2250$. $k=6$: $2700$. None of these consist only of $0$'s and $1$'s.

    \textcolor{actsgray}{\textit{[$\sim$3k tokens: continues enumerating $450 \cdot k$ for $k = 7, \ldots, 100$. None match the constraint.]}}

    Hmm, okay, so even up to $k = 100$, we haven't found a number with only $0$'s and $1$'s. So perhaps $11111111100$ is indeed the minimal.

    \textcolor{actsred}{But wait, maybe another $11$-digit number is smaller.} Let me try $10111111100$. It ends with $00$ \checkmark. Digit sum: $1 + 0 + 1 + 1 + 1 + 1 + 1 + 1 + 1 + 0 + 0 = 9$ \checkmark. 
    
    So $10111111100$ is also a multiple of $450$, and $10{,}111{,}111{,}100 < 11{,}111{,}111{,}100$.

    Wait, maybe we can rearrange further. Try $10011111100$: digit sum $1+0+0+1+1+1+1+1+1+0+0 = 8$. \textcolor{actsred}{$8$ is not a multiple of $9$,} so not divisible. $10101111100$: digit sum $8$. $10110111100$: digit sum $8$. Still not.

    \textcolor{actsgray}{\textit{[$\sim$3k tokens: enumerates further $11$-digit rearrangements with a single $0$ shifted in among the $1$'s; each one has digit sum $8$ and is invalid. Also checks $12$-digit candidates like $101111101100$ (sum $9$ but $12$ digits, so larger).]}}

    So none of these have a digit sum of $9$, except for $11111111100$ and $10111111100$. Therefore, $10111111100$ is the only $11$-digit candidate smaller than $11111111100$.

    \textcolor{actsgray}{\textit{[$\sim$2.5k tokens: checks that no number between $10{,}000{,}000{,}000$ and $10{,}111{,}111{,}100$ meets the criteria, by examining several intermediate multiples of $450$.]}}

    Divided by $450$: $10{,}111{,}111{,}100 / 450 = 22{,}469{,}136$. Verify: $450 \times 22{,}469{,}136 = 10{,}111{,}111{,}100$. \checkmark

    Therefore, the minimal number is $\boxed{10111111100}$.

    \textcolor{actsgray}{\texttt{</think>}}

    To find the smallest positive multiple of $450$ whose digits are all zeroes and ones, we need to consider the divisibility rules for $450$\ldots\ $\boxed{10111111100}$.

    \vspace{4pt}
    \textcolor{actsred}{\textbf{Wrong} ($11{,}178$ tokens). The digit sum of $10111111100$ is actually $8$ (one~$1$, then $0$, then seven~$1$'s, then two~$0$'s, totalling eight $1$'s), not $9$. Vanilla miscounted, and the division check $450 \!\times\! 22{,}469{,}136 \!=\! 10{,}111{,}111{,}200 \!\ne\! 10{,}111{,}111{,}100$ silently confirmed the wrong number.}
    \end{minipage}\hfill
    \begin{minipage}[t]{0.475\textwidth}
    \scriptsize
    \setlength{\parskip}{3pt}
    \begin{center}{\small\textbf{ACTS}}\end{center}

    \textcolor{actsgray}{\texttt{<think>}}

    \textcolor{actsorange}{Okay, so I need to find} the smallest positive multiple of $450$ where all the digits are either $0$ or $1$. Hmm, that sounds interesting. Let me break this down and think through it step by step.

    \textcolor{actsorange}{First, let's recall that} a multiple of $450$ must be divisible by both $450$ and, well, $450$ itself. But actually, since $450 = 45 \times 10 = 9 \times 5 \times 10 = 2 \times 3^2 \times 5^2$, for a number to be divisible by $450$ it must be divisible by $2$, $3^2$, and $5^2$.

    \textcolor{actsorange}{Divisible by} $2$: that means the number must be even, so its last digit has to be $0$. Since all digits are $0$ or $1$, that's already satisfied because the only even digit available is $0$. So the number must end with a $0$.

    \textcolor{actsorange}{Divisible by} $5^2 = 25$: for a number to be divisible by $25$, its last two digits must be $00$, $25$, $50$, or $75$. But our digits are only $0$ or $1$, so the only possible ending is $00$. So the number must end with two $0$'s.

    \textcolor{actsorange}{Divisible by} $3^2 = 9$: a number is divisible by $9$ if the sum of its digits is a multiple of $9$. Since our number consists only of $1$'s and $0$'s, the digit sum is just the number of $1$'s. So the count of $1$'s must be a multiple of $9$.

    \textcolor{actsorange}{Putting that all} together: the number is composed only of $1$'s and $0$'s, ends with two $0$'s, and the count of $1$'s is a multiple of $9$.

    \textcolor{actsorange}{So, the problem} reduces to finding the smallest number composed of $1$'s and $0$'s, ending with ``$00$'', with exactly nine $1$'s (since $9$ is the smallest positive multiple of $9$).

    \textcolor{actsorange}{But wait, actually, the number could have} more than two trailing $0$'s, but the minimal one will have exactly two. So we need a number with nine $1$'s and ending with two $0$'s, arranged so as to make the number as small as possible.

    \textcolor{actsorange}{To make} the number as small as possible, the $1$'s should be packed as far to the right as possible, so the digit count is minimized.

    \textcolor{actsorange}{Wait, no.} The first digit has to be $1$ since the number is positive, so the structure is $1$ followed by some combination of $1$'s and $0$'s, ending with two $0$'s. But to make $N$ as small as possible, we still want the $1$'s as far right as possible.

    \textcolor{actsorange}{Wait, perhaps we can have} multiple $1$'s spread out but in such a way that the number is minimized.

    \textcolor{actsorange}{Alternatively, maybe the} minimal number is the smallest multiple of $450$ that consists only of $0$'s and $1$'s, which would likely have many leading $1$'s and end with $00$. But perhaps there's a smarter way to approach this.

    \textcolor{actsorange}{Since the number} must be divisible by $25$, write $N = 100 \cdot k$ with $k$ composed only of $0$'s and $1$'s. Since the trailing two $0$'s contribute nothing to the digit sum, $9 \mid N$ iff $9 \mid k$, so $k$ must have nine $1$'s.

    \textcolor{actsorange}{Therefore, the sum} of digits of $k$ is $9$, i.e., $k$ has exactly nine $1$'s and any number of $0$'s. Therefore, the problem reduces to finding the smallest $k$ composed of nine $1$'s and any number of $0$'s.

    \textcolor{actsorange}{Therefore, the} minimal $N$ is the minimal $k$ multiplied by $100$. Therefore, our task is to find the smallest $k$ which is a number with exactly nine $1$'s and the rest $0$'s.

    \textcolor{actsorange}{So, how do} we find the minimal $k$? To find the smallest number with nine $1$'s, the number should have the smallest digit length. With nine $1$'s and no extra $0$'s, $k = 111111111$ ($9$ digits).

    \textcolor{actsorange}{Wait, actually,} the smallest number with nine $1$'s would just be $111111111$; adding any $0$ only makes $k$ longer and therefore larger.

    \textcolor{actsorange}{Wait, no.} For example $1000000000$ is smaller than $111111111$, but $k$ must contain nine $1$'s; with that constraint $111111111$ is indeed the smallest valid $k$. The smallest $N$ is therefore $111111111 \times 100 = 11{,}111{,}111{,}100$.

    \textcolor{actsorange}{\textbf{Final Answer.}} $\boxed{11{,}111{,}111{,}100}$.

    \textcolor{actsgray}{\texttt{</think>}}

    To find the smallest positive multiple of $450$ whose digits are all zeroes and ones, we need to consider the following\ldots\ $\boxed{11{,}111{,}111{,}100}$.

    \vspace{4pt}
    \textcolor{actsorange}{\textbf{Correct} ($1{,}948$ tokens).}
    \end{minipage}
    \end{tcolorbox}
    \caption{\textbf{Rescue.} Vanilla finds $11{,}111{,}111{,}100$ early but then commits to $10{,}111{,}111{,}100$ after miscounting its digit sum as $9$ (it is $8$). ACTS reaches the right candidate via structured stepping and concludes correctly.}
    \label{fig:case-rescue}
\end{figure*}

\begin{figure*}[!t]
    \definecolor{promptback}{RGB}{252, 252, 252}
    \definecolor{promptframe}{RGB}{138, 147, 156}
    \definecolor{actsorange}{RGB}{210, 120, 30}
    \definecolor{actsred}{RGB}{170, 60, 60}
    \definecolor{actsgray}{RGB}{100, 100, 100}

    \begin{tcolorbox}[
      enhanced jigsaw,
      colback=promptback,
      colframe=promptframe,
      coltitle=white,
      fonttitle=\bfseries,
      arc=2pt,
      boxrule=0.5pt,
      title={Case 2: Shorten},
      overlay={
        \draw[gray!55, line width=0.4pt, dashed]
          ([yshift=-4pt]interior.north) -- ([yshift=4pt]interior.south);
      },
    ]
    \noindent
    {\small\textbf{Problem.} What is the base six equivalent of $999_{10}$?\\\textbf{Ground truth:} $4343_6$}
    \vspace{2pt}

    \noindent
    \begin{minipage}[t]{0.475\textwidth}
    \scriptsize
    \setlength{\parskip}{3pt}
    \begin{center}{\small\textbf{Vanilla}}\end{center}

    \textcolor{actsgray}{\texttt{<think>}}

    Alright, so I have to figure out how to convert the decimal number $999$ into base six. Base conversions can be tricky, but I remember the method involves dividing the number by the target base and tracking the remainders. Each remainder becomes a digit in the new base, starting from the least significant.

    Let me write this down step by step. $999 \div 6$: $6 \times 166 = 996$, so quotient $166$, remainder $3$. $166 \div 6$: $6 \times 27 = 162$, so quotient $27$, remainder $4$. $27 \div 6$: $6 \times 4 = 24$, so quotient $4$, remainder $3$. $4 \div 6$: $0$ quotient, remainder $4$.

    So the remainders from first to last are $3, 4, 3, 4$. Writing them in reverse order gives the base-six number $4343$.

    Hmm, let me cross-check that to be sure. Convert $4343_6$ back to decimal: $4 \cdot 6^3 + 3 \cdot 6^2 + 4 \cdot 6 + 3 = 4 \cdot 216 + 3 \cdot 36 + 4 \cdot 6 + 3 = 864 + 108 + 24 + 3 = 999$. Perfect, that matches.

    \textcolor{actsred}{Wait, but hold on,} when I was converting initially, I thought the first remainder was the least significant digit\ldots\ let me recount the steps to make sure. Yes, the first remainder $3$ is the units digit; the last remainder $4$ is the most significant. Reading from last to first gives $4343$. OK, that's consistent.

    Just to be thorough, let me verify each division step once more. $999 \div 6$: $6 \cdot 166 = 996$, remainder $3$. $166 \div 6$: $6 \cdot 27 = 162$, remainder $4$. $27 \div 6$: $6 \cdot 4 = 24$, remainder $3$. $4 \div 6$: $0$ quotient, remainder $4$. So the remainders are $3, 4, 3, 4$, and reading in reverse gives $4343$. Consistent.

    \textcolor{actsred}{Alternatively, another way is to find the largest power of $6$ that fits into $999$.} Powers of $6$: $6^0 \!=\! 1$, $6^1 \!=\! 6$, $6^2 \!=\! 36$, $6^3 \!=\! 216$, $6^4 \!=\! 1296 > 999$. So the highest power we need is $6^3$.

    $999 \div 216 = 4$ rem $135$ (coefficient of $6^3$ is $4$). $135 \div 36 = 3$ rem $27$ (coefficient of $6^2$ is $3$). $27 \div 6 = 4$ rem $3$ (coefficient of $6^1$ is $4$). Final coefficient of $6^0$ is $3$. Putting it all together: $4343_6$.

    This matches the previous result. \textcolor{actsred}{So that's a good confirmation.}

    \textcolor{actsgray}{\textit{[$\sim$1.5k tokens: writes out the division steps once more in tabular form; explores Horner's-method angle as another cross-check.]}}

    \textcolor{actsred}{Alternatively, someone might try to convert $999$ into binary first and then convert that binary number into base six.} Let me try that approach.

    $999 = 512 + 256 + 128 + 64 + 32 + 4 + 2 + 1$, so $999_{10} = 1111100111_2$. Verifying: $512 + 256 + 128 + 64 + 32 + 4 + 2 + 1 = 999$. \checkmark

    \textcolor{actsgray}{\textit{[$\sim$2.5k tokens: attempts to convert the binary representation into base $6$ via grouping; the grouping doesn't yield a clean partition, so it re-derives $4343$ from scratch a couple more times.]}}

    Thus, the base six equivalent of $999_{10}$ is $\boxed{4343_6}$.

    \textcolor{actsgray}{\texttt{</think>}}

    To convert $999_{10}$ to base six, we use the method of dividing by $6$ and tracking the remainders\ldots\ $\boxed{4343_6}$.

    \vspace{4pt}
    \textcolor{actsred}{\textbf{Correct} ($6{,}346$ tokens), but $\sim$$4/5$ of those tokens were spent re-deriving the same answer through alternative methods.}
    \end{minipage}\hfill
    \begin{minipage}[t]{0.475\textwidth}
    \scriptsize
    \setlength{\parskip}{3pt}
    \begin{center}{\small\textbf{ACTS}}\end{center}

    \textcolor{actsgray}{\texttt{<think>}}

    \textcolor{actsorange}{Okay, so I need to figure out what} $999$ in base ten is when converted to base six. Hmm, base conversions can be tricky, but I think I remember some steps. I think the method involves dividing the number by the base and keeping track of the remainders. So, since we're converting to base six, I should divide $999$ by $6$ repeatedly and record the remainders each time. Then, the base six number is the remainders read from bottom to top.

    \textcolor{actsorange}{Wait, let me make sure.} The process is: (1) divide the number by $6$; (2) write down the remainder; (3) replace the number with the quotient from the division; (4) repeat steps (1)--(3) until the quotient is $0$; (5) the base six number is the sequence of remainders read in reverse order.

    \textcolor{actsorange}{Okay, let me try that.} First, I'll divide $999$ by $6$. $999$ divided by $6$: $6$ goes into $999$ how many times? Well, $6 \times 166 = 996$. So the quotient is $166$ and the remainder is $999 - 996 = 3$. I write down $3$.

    \textcolor{actsorange}{Now,} take the quotient $166$ and divide by $6$. $166$ divided by $6$: $6 \times 27 = 162$, so the quotient is $27$ and the remainder is $166 - 162 = 4$. I write down $4$.

    \textcolor{actsorange}{Next,} take the quotient $27$ and divide by $6$. $27$ divided by $6$ is $4$ with a remainder of $3$, because $6 \times 4 = 24$ and $27 - 24 = 3$. I write down $3$.

    \textcolor{actsorange}{Now,} take the quotient $4$ and divide by $6$. $4$ divided by $6$: that's less than $6$, so the quotient is $0$ and the remainder is $4$. I write down $4$.

    \textcolor{actsorange}{Now, since the} quotient is $0$, I stop here.

    \textcolor{actsorange}{So, the} remainders I got, in the order they were obtained, are $3, 4, 3, 4$.

    \textcolor{actsorange}{But wait,} in the conversion process, you read the remainders from the last division to the first. The first remainder is the least significant digit; the last remainder is the most significant.

    \textcolor{actsorange}{Wait, let me check.} The first division gave quotient $166$ and remainder $3$ -- that $3$ is the units place in base six.

    \textcolor{actsorange}{Then,} dividing $166$ by $6$ gave a remainder of $4$, which is the sixes place.

    \textcolor{actsorange}{Then,} dividing $27$ by $6$ gave a remainder of $3$, which is the thirty-sixes place.

    \textcolor{actsorange}{Then,} dividing $4$ by $6$ gave a remainder of $4$, which is the two-hundred-sixteens place.

    \textcolor{actsorange}{So,} reading the remainders from last to first, that would be $4, 3, 4, 3$.

    \textcolor{actsorange}{Therefore,} $999$ in base six is $4343$.

    \textcolor{actsorange}{Wait, let me verify that.} Maybe I can check by converting $4343$ from base six back to base ten and see if it equals $999$. So $4343_6 = 4 \cdot 6^3 + 3 \cdot 6^2 + 4 \cdot 6^1 + 3 \cdot 6^0 = 864 + 108 + 24 + 3 = 999$. \checkmark

    \textcolor{actsorange}{So,} the conversion is correct.

    \textcolor{actsorange}{\textbf{Final Answer.}} The base six equivalent of $999_{10}$ is $\boxed{4343_6}$.

    \textcolor{actsgray}{\texttt{</think>}}

    To convert $999_{10}$ to base six, we repeatedly divide the number by $6$ and record the remainders\ldots\ $\boxed{4343_6}$.

    \vspace{4pt}
    \textcolor{actsorange}{\textbf{Correct} ($1{,}053$ tokens).}
    \end{minipage}
    \end{tcolorbox}
    \caption{\textbf{Shorten.} Vanilla derives $4343$ early then re-derives it via powers of $6$ and a binary intermediate. ACTS does the division once, verifies once, and concludes.}
    \label{fig:case-shorten}
\end{figure*}

\end{document}